\documentclass[11pt,a4paper]{article}
\usepackage[hyperref]{acl2017}
\usepackage{times}
\usepackage{latexsym}

\usepackage{url}

\newcommand{\eat}[1]{\ignorespaces}

\usepackage{verbatim}
\usepackage{amsmath}
\usepackage{bm}
\usepackage{booktabs}

\newcommand{\Nor}{\mathcal{N}}
\usepackage{amsthm}

\usepackage{times}
\usepackage{graphicx} 
\usepackage{subfigure} 

\usepackage{natbib}

\defcitealias{tensorflow}{Abadi et. al, 2015}

\usepackage{algorithm}
\usepackage{algorithmic}

\usepackage{float}

\aclfinalcopy 
\def\aclpaperid{145} 

\title{Multimodal Word Distributions}

\author{Ben Athiwaratkun \\
 Cornell University \\
  {\tt pa338@cornell.edu} \\\And
 Andrew Gordon Wilson\\
  Cornell University \\
  {\tt andrew@cornell.edu} \\}

\date{}

\begin{document}
\maketitle

\begin{abstract}
Word embeddings provide point representations of 
words containing useful semantic information. 
We introduce multimodal word distributions formed
from Gaussian mixtures, for 
multiple word meanings, entailment, and rich 
uncertainty information.  To learn these distributions,
we propose an energy-based max-margin objective.
We show that the resulting approach captures uniquely 
expressive semantic information, and 
outperforms alternatives, such as word2vec
skip-grams, and Gaussian embeddings, on 
benchmark datasets such as word similarity and entailment. 
\end{abstract}

\section{Introduction} \label{section:introduction}
To model language, we must represent words.  We can imagine representing every word with a binary one-hot vector corresponding to a dictionary position.  But such a representation contains no valuable semantic information: distances between word vectors represent only differences in alphabetic ordering.  Modern approaches, by contrast, learn to map words with similar meanings to nearby points in a vector space \citep{word2vec2}, from large datasets such as Wikipedia.  These learned word embeddings have become ubiquitous in predictive tasks.

\citet{word2gauss} recently proposed an alternative view, where words are represented by a whole probability distribution instead of a deterministic point vector.  Specifically, they model each word by a Gaussian distribution, and learn its mean and covariance matrix from data.  This approach generalizes any deterministic point embedding, which can be fully captured by the mean vector of the Gaussian distribution.  Moreover, the full distribution provides much richer information than point estimates for characterizing words, representing probability mass and uncertainty across a set of semantics.

However, since a Gaussian distribution can have only one mode, the learned uncertainty in this representation can be overly diffuse for words with multiple distinct meanings (polysemies), in order for the model to assign \emph{some} density to any plausible semantics \citep{word2gauss}.  
Moreover, the mean of the Gaussian can be pulled in many opposing directions, leading to a biased distribution that centers its mass mostly around one meaning while leaving the others not well represented.

In this paper, we propose to represent each word with an expressive multimodal distribution, for multiple distinct meanings, entailment, heavy tailed uncertainty, and enhanced interpretability.  For example, one mode of the word `bank' could overlap with distributions for words such as `finance' and `money', and another mode could overlap with the distributions for `river' and `creek'. It is our contention that such flexibility is critical for both qualitatively learning about the meanings of words, and for optimal performance on many predictive tasks.

In particular, we model each word with a mixture of Gaussians (Section~\ref{subsection:wordrep}).  We learn all the parameters of this mixture model using a maximum margin energy-based ranking objective \citep{maxmargin_thorsten, word2gauss} (Section~\ref{subsection:objective}), where the energy function describes the affinity between a pair of words.  For analytic tractability with Gaussian mixtures, we use the inner product between probability distributions in a Hilbert space, known as the expected likelihood kernel \citep{prob_product_kernel}, as our energy function (Section~\ref{subsection:energy}).  Additionally, we propose transformations for numerical stability and initialization~\ref{sup:hyperparams}, resulting in a robust, straightforward, and scalable learning procedure, capable of training on a corpus with billions of words in days.  
We show that the model is able to automatically discover multiple meanings for words (Section~\ref{subsection:qualitative_eval}), and 
significantly outperform other alternative methods across several tasks such as word similarity and entailment (Section~\ref{subsection:wordsim},~\ref{subsection:wordsim2},~\ref{subsection:entailment}).  We have made code available at \url{http://github.com/benathi/word2gm}, where we implement our model in Tensorflow \citepalias{tensorflow}.

\section{Related Work} \label{section:relatedwork}

In the past decade, there has been an explosion of interest in word vector representations. {\tt word2vec}, arguably the most popular word embedding, uses continuous bag of words and skip-gram models, in conjunction with negative sampling for efficient conditional probability estimation \cite{word2vec2, word2vec1}.
Other popular approaches use feedforward \citep{nnlm} and recurrent neural network language models \citep{rnnlm1, rnnlm2, collobert_we} to predict missing words in sentences, producing hidden layers that can act as word embeddings that encode semantic information. 
They employ  conditional probability estimation techniques, including hierarchical softmax \cite{DBLP:conf/asru/MikolovDPBC11, DBLP:conf/nips/MnihH08, DBLP:conf/aistats/MorinB05} and noise contrastive estimation \cite{nce}.

A different approach to learning word embeddings is through factorization of word co-occurrence matrices such as  {\tt GloVe} embeddings \cite{glove}.
 The matrix factorization approach has been shown to have an implicit connection with skip-gram and negative sampling \citet{neuralemb_matrixfac}. Bayesian matrix factorization where  row and columns are modeled as Gaussians has been explored in \citet{bayesian_pmf} and provides a different probabilistic perspective of word embeddings.

In exciting recent work, \citet{word2gauss} propose a Gaussian distribution to model each word.  Their approach is significantly more expressive than typical point embeddings, with the ability to represent concepts such as \emph{entailment}, by having the distribution for one word (e.g. `music') encompass the distributions for sets of related words (`jazz' and `pop'). However, with a unimodal distribution, their approach cannot capture multiple distinct meanings, much like most deterministic approaches. 

Recent work has also proposed deterministic embeddings that can capture polysemies, for example through a cluster centroid of context vectors \citep{multipleprototypes}, or an adapted skip-gram model with an EM algorithm to learn multiple latent representations per word \citep{multi_word_embs}.
\citet{nonparam_multiprototype} also extends skip-gram with multiple prototype embeddings where the number of senses per word is determined by a non-parametric approach. \citet{topical_word_emb} learns topical embeddings based on latent topic models where each word is associated with multiple topics.  Another related work by \citet{infinite_word_embs} models embeddings in infinite-dimensional space where each embedding can gradually represent incremental word sense if complex meanings are observed. Although independent of our work, we later found that \citet{gm_multiple_prototypes} proposed a similar model to ours; however, our setup obtains significantly improved results on all evaluation metrics.

Probabilistic word embeddings have only recently begun to be explored, and have so far shown great promise.  In this paper, we propose probabilistic word embedding that can capture multiple meanings.
We use a Gaussian mixture model which allows for a highly expressive distributions over words.  At the same time, we retain scalability and analytic tractability with an expected likelihood kernel energy function for training.  
The model and training procedure harmonize to learn descriptive representations of words, with 
superior performance on several benchmarks.

\section{Methodology} \label{section:methodology}

In this section, we introduce our Gaussian mixture (GM) model for word representations, and present a training method to learn the parameters of the Gaussian mixture.  This method uses an energy-based maximum margin objective, where we wish to maximize the similarity of distributions of nearby words in sentences.  We
propose an energy function that compliments the GM model by retaining analytic tractability.  We also provide critical practical details for numerical stability, hyperparameters, and initialization. 

\subsection{Word Representation} \label{subsection:wordrep}
We represent each word $w$ in a dictionary as a Gaussian mixture with $K$ components.  Specifically, the distribution of $w$, 
$f_w$, is given by the density
\begin{align}
f_w(\vec{x}) &= \sum_{i=1}^K p_{w,i} \ \Nor  \left[ \vec{x}; \vec{\mu}_{w,i} , \Sigma_{w,i}  \right] \label{eqn: mix} \\
&= \sum_{i=1}^K  \frac{p_{w,i} }{\sqrt{2 \pi | \Sigma_{w,i} | }} e^{-\frac{1}{2} (\vec{x} - \vec{\mu}_{w,i})^{\top} \Sigma_{w,i}^{-1} (\vec{x} - \vec{\mu}_{w,i})} \,, \notag
\end{align} 
where $\sum_{i=1}^K p_{w,i} = 1$. 

\begin{figure}[H]
\begin{center}
\centerline{\includegraphics[width=\columnwidth,trim={220 260 160 215},clip]{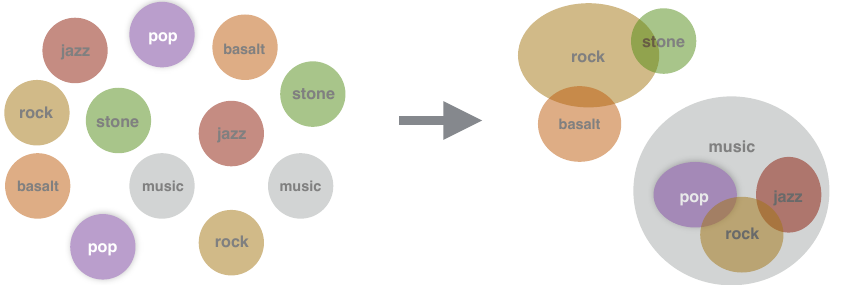}}
\centerline{\includegraphics[width=\columnwidth,trim={220 262 160 210},clip]{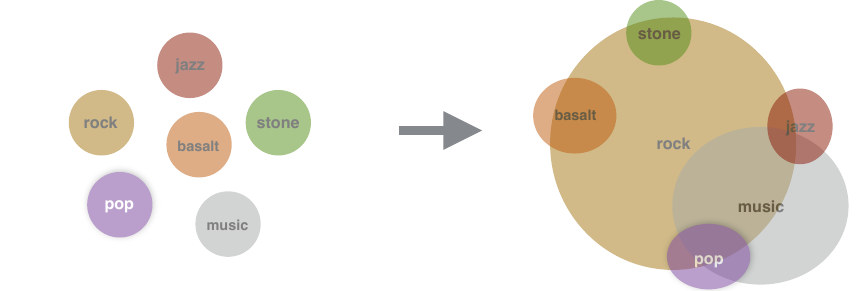}}
\caption{
\textbf{Top:} A Gaussian Mixture embedding, where each component corresponds to a distinct meaning.
Each Gaussian component is represented by an ellipsoid, whose center is specified by the mean vector and contour surface specified by the covariance matrix, reflecting subtleties in meaning and uncertainty.
On the left, we show examples of Gaussian mixture distributions of words where Gaussian components are randomly initialized. After training, we see on the right that one component of the word  `rock' is closer to `stone' and `basalt', whereas the other component is closer to `jazz' and `pop'. 
We also demonstrate the entailment concept where the distribution of the more general word `music' encapsulates words such as `jazz', `rock', `pop'. 
\textbf{Bottom:} A Gaussian embedding model \cite{word2gauss}.
For words with multiple meanings, such as `rock', the variance of the learned representation becomes unnecessarily large in order to assign some probability to both meanings.  Moreover, the mean vector for such words can be pulled between two clusters, centering the mass of the distribution on a region which is far from certain meanings.
}
\label{fig:multimodal}
\end{center}
\end{figure}

The mean vectors $\vec{\mu}_{w,i}$ represent the location of the $i^{th}$ component of word $w$, and are akin to the point embeddings provided by popular approaches like \texttt{word2vec}. $p_{w,i}$ represents the component probability (mixture weight), and $\Sigma_{w,i}$ is the component covariance matrix, containing uncertainty information.  Our goal is to learn all of the model parameters $\vec{\mu}_{w,i}, p_{w,i}, \Sigma_{w,i}$ from a corpus of natural sentences to extract semantic information of words. Each Gaussian component's mean vector of word $w$ can represent one of the word's distinct meanings. For instance, one component of a polysemous word such as `rock' should represent the meaning related to `stone' or `pebbles', whereas another component should represent the meaning related to music such as `jazz' or `pop'. Figure ~\ref{fig:multimodal} illustrates our word embedding model, and the difference between multimodal and unimodal representations, for words with multiple meanings.

\subsection{Skip-Gram} 
The training objective for learning $\theta = \{\vec{\mu}_{w,i}, p_{w,i}, \Sigma_{w,i}\}$ draws inspiration from the continuous skip-gram model \cite{word2vec2}, where word embeddings are trained to maximize the probability of observing a word given another nearby word. This procedure follows the \emph{distributional hypothesis} that words occurring in natural contexts tend to be semantically related.  For instance, the words `jazz'  and `music' tend to occur near one another more often than `jazz' and `cat'; hence, `jazz' and `music' are more likely to be related. The learned word representation contains useful semantic information and can be used to perform a variety of NLP tasks such as word similarity analysis, sentiment classification, modelling word analogies, or as a preprocessed input for complex system such as statistical machine translation.

\subsection{Energy-based Max-Margin Objective} \label{subsection:objective}
\label{sec: maxmarg}
Each sample in the objective consists of two pairs of words, $(w,c)$ and $(w,c')$. $w$ is  sampled from a sentence in a corpus and $c$ is a nearby word within a context window of length $\ell$. For instance, a word $w = $ `jazz' which occurs in the sentence `I listen to jazz music' has context words (`I', `listen', `to' , `music'). $c'$ is a negative context word (e.g. `airplane') obtained from random sampling. 

The objective is to maximize the energy between words that occur near each other, $w$ and $c$, and minimize the energy between $w$ and its negative context $c'$. This approach is similar to negative sampling \citep{word2vec2, word2vec1}, which contrasts the dot product between positive context pairs with negative context pairs. The energy function is a measure of similarity between distributions and will be discussed in Section~\ref{subsection:energy}. 

We use a max-margin ranking objective \cite{maxmargin_thorsten}, used for Gaussian embeddings in \citet{word2gauss}, which pushes the similarity of a word and its positive context higher than that of its negative context by a margin $m$:
\begin{align}
\nonumber L_\theta (w, c, c') = \max(0, \ \ \ \ \ \ \ \ \ \ \ \ \ \ \ \ \ \ \ \ \ \ \ \ \ \ \ \ \ \ \ \ \ \ \ \ \ \ \ \ \\
\nonumber m - \log E_\theta(w, c)  + \log E_\theta(w, c') ) 
\end{align}
This objective can be minimized by mini-batch stochastic gradient descent with respect to the parameters $\theta = \{\vec{\mu}_{w,i}, p_{w,i}, \Sigma_{w,i}\}$ -- the mean vectors, covariance matrices, and mixture weights -- of our multimodal embedding in Eq.~\eqref{eqn: mix}. 

\paragraph{Word Sampling}
We use a word sampling scheme similar to the implementation in {\tt word2vec} \cite{word2vec2, word2vec1} to balance the importance of frequent words and rare words. Frequent words such as  `the', `a', `to' are not as meaningful as relatively less frequent words such as `dog', `love', `rock', and we are often more interested in learning the semantics of the less frequently observed words. We use subsampling to improve the performance of learning word vectors \cite{word2vec1}. This technique discards word $w_i$ with probability $P(w_i) = 1 - \sqrt{t/f(w_i)}$, where $f(w_i)$ is the frequency of word $w_i$ in the training corpus and $t$ is a frequency threshold.  

To generate negative context words, each word type $w_i$ is sampled according to a distribution $P_n(w_i) \propto U(w_i)^{3/4}$ which is a distorted version of the unigram distribution $U(w_i)$ that also serves to diminish the relative importance of frequent words.  Both subsampling and the negative distribution choice are proven  effective in {\tt word2vec} training \citep{word2vec1}.

\subsection{Energy Function}
\label{subsection:energy}

For vector representations of words, a usual choice for similarity measure (energy function) is a dot product between two vectors. Our word representations are distributions instead of point vectors and therefore need a measure that reflects not only the point similarity, but also the uncertainty. 

\subsubsection{Expected Likelihood Kernel} \label{subsection:elk}
We propose to use the \emph{expected likelihood kernel}, which is a generalization of an inner product between vectors to an inner product between distributions \cite{prob_product_kernel}. 
That is, 
\[
E(f,g) =  \int f(x) g(x) \ d x =  \langle f, g \rangle_{L_2} 
\]
where $\langle \cdot, \cdot \rangle_{L_2} $ denotes the inner product in Hilbert space $L_2$. We choose this form of energy since it  can be evaluated in a closed form given our choice of probabilistic embedding in Eq.~\eqref{eqn: mix}.

For Gaussian mixtures $f,g$ representing the words $w_f, w_g$,  $f(x) = \sum_{i=1}^K p_i \Nor(x; \vec{\mu}_{f,i} , \Sigma_{f,i} ) $ and $g(x)  =  \sum_{i=1}^K q_i \Nor(x; \vec{\mu}_{g,i} , \Sigma_{g,i} )$, $\sum_{i =1}^K p_i = 1 $, and $\sum_{i =1}^K q_i = 1$, we find (see Section~\ref{subsection:derivation}) the log energy is 
\begin{equation} \label{eq:loge}
 \log E_\theta(f,g) = \log \sum_{j=1}^K \sum_{i=1}^K p_i q_j e^{\xi_{i,j}} 
\end{equation}
where
\begin{align}
\nonumber
\xi_{i,j} &\equiv \log  \Nor(0; \vec{\mu}_{f,i} - \vec{\mu}_{g,j}, \Sigma_{f,i} + \Sigma_{g,j} ) \\ \nonumber
&= - \frac{1}{2} \log \det( \Sigma_{f,i} + \Sigma_{g,j} ) - \frac{D}{2} \log (2 \pi)  \\ 
  - \frac{1}{2} & (\vec{\mu}_{f,i} - \vec{\mu}_{g,j} )^\top (\Sigma_{f,i} + \Sigma_{g,j} )^{-1} (\vec{\mu}_{f,i} - \vec{\mu}_{g,j} )  \label{eq:partial_energy}
 \end{align}
We call the term $\xi_{i,j}$ partial (log) energy. Observe that this term captures the similarity between the $i^{th}$ meaning of word $w_f$ and the $j^{th}$ meaning of word $w_g$. The total energy in Equation~\ref{eq:loge}  is the  sum of possible pairs of partial energies, weighted accordingly by  the mixture probabilities $p_i$ and $q_j$.
 
The term $- (\vec{\mu}_{f,i} - \vec{\mu}_{g,j} )^\top (\Sigma_{f,i} + \Sigma_{g,j} )^{-1} (\vec{\mu}_{f,i} - \vec{\mu}_{g,j} ) $ in  $\xi_{i,j}$ explains the difference in mean vectors of semantic pair $(w_f, i)$ and $(w_g, j)$. If the semantic uncertainty (covariance) for both pairs are low, this term has more importance relative to other terms due to the inverse covariance scaling. We observe that the loss function $L_\theta$ in Section~\ref{sec: maxmarg} attains a low value when $E_\theta(w,c)$ is relatively high. High values of $E_\theta(w,c)$ can be achieved when the component means across different words $\vec{\mu}_{f,i}$ and $\vec{\mu}_{g,j}$ are close together (e.g., similar point representations).  High energy can also be achieved by large values of $\Sigma_{f,i}$ and $\Sigma_{g,j}$, which washes out the importance of the mean vector difference. The term $- \log \det( \Sigma_{f,i} + \Sigma_{g,j} )$ serves as a regularizer that prevents the covariances from being pushed too high at the expense of learning a good mean embedding.

At the beginning of training, $\xi_{i,j}$ roughly are on the same scale among all pairs $(i,j)$'s. During this time, all components learn the signals from the word occurrences equally. As training progresses and the semantic representation of each mixture becomes more clear, there can be one term of  $\xi_{i,j}$'s that is predominantly higher than other terms, giving rise to a semantic pair that is most related. 

\subsubsection{Probability Product Kernel} \label{subsection:ppk}  

In general, the probability product kernel $K_\rho(f,g) = \int f(x)^\rho g(x)^\rho \ dx $ for $\rho >0$ between two Gaussians are:
\begin{align*}
\xi_{i,j}^\rho \equiv  \log K_\rho(f_i,g_j)   \\
= (1-2 \rho) \frac{D}{2} \log (2 \pi)  - \frac{D}{2} \log(\rho)  \\
 + \log \det \left[ \Sigma_{f,i}^{\rho-1} \Sigma_{g,j}^\rho + \Sigma_{f,i}^{\rho} \Sigma_{g,j}^{\rho-1}  \right]  \\ 
 - \frac{\rho}{2} (\mu_{f,i} - \mu_{g,j} ) (\Sigma_{f,i} + \Sigma_{g,j})^{-1} (\mu_{f,i} - \mu_{g,j} )  
\end{align*}
For mixture of Gaussians, we have 
\[
\log E_\theta^\rho (f,g) = \sum_{i=1}^{K} \sum_{j=1}^{K} (p_i q_j)^\rho e^{\xi_{i,j}}
\]

Note that for the case where $\rho =1$, we recover the expected likelihood kernel in Section \ref{subsection:elk}

\subsubsection{Other Energy Functions}
The negative KL divergence is another sensible choice of energy function, providing an asymmetric metric between word distributions. However, unlike the expected likelihood kernel, KL divergence does not have a closed form if the two distributions are Gaussian mixtures.

\begin{table*}[h]
\begin{center}
\begin{small}
\begin{tabular}{ccc}
\toprule
Word & Co. & Nearest Neighbors \\
\midrule
rock	& 0 	& basalt:1, boulder:1, boulders:0, stalagmites:0, stalactites:0, rocks:1, sand:0, quartzite:1, bedrock:0 \\
rock 	& 1 	& rock/:1, ska:0, funk:1, pop-rock:1, punk:1, indie-rock:0, band:0, indie:0, pop:1 \\
bank & 0 & banks:1, mouth:1, river:1, River:0, confluence:0, waterway:1, downstream:1, upstream:0, dammed:0 \\
bank & 1 & banks:0, banking:1, banker:0, Banks:1, bankas:1, Citibank:1, Interbank:1, Bankers:0, transactions:1 \\
Apple & 0 & Strawberry:0, Tomato:1, Raspberry:1, Blackberry:1, Apples:0, Pineapple:1, Grape:1, Lemon:0
\\
Apple & 1 & Macintosh:1, Mac:1, OS:1, Amiga:0, Compaq:0, Atari:1, PC:1, Windows:0, iMac:0  
\\
star & 0 & stars:0, Quaid:0, starlet:0, Dafoe:0, Stallone:0, Geena:0, Niro:0, Zeta-Jones:1, superstar:0 \\
star & 1 & stars:1, brightest:0, Milky:0, constellation:1, stellar:0, nebula:1, galactic:1, supernova:1, Ophiuchus:1 \\
cell & 0 & cellular:0, Nextel:0, 2-line:0, Sprint:0, phones.:1, pda:1, handset:0, handsets:1, pushbuttons:0 \\
cell & 1 & cytoplasm:0, vesicle:0, cytoplasmic:1, macrophages:0, secreted:1, membrane:0, mitotic:0, endocytosis:1
\\
left & 0 & After:1, back:0, finally:1, eventually:0, broke:0, joined:1, returned:1, after:1, soon:0 \\
left & 1  & right-hand:0, hand:0, right:0, left-hand:0, lefthand:0, arrow:0, turn:0, righthand:0, Left:0 \\
\bottomrule
\end{tabular}

\smallskip
\begin{tabular}{ccc}
\toprule
Word  & Nearest Neighbors \\
\midrule
rock & 
band, bands, Rock, indie, Stones, breakbeat, punk, electronica, funk \\
bank & banks, banking, trader, trading, Bank, capital, Banco, bankers, cash \\
Apple & Macintosh, Microsoft, Windows, Macs, Lite, Intel, Desktop, WordPerfect, Mac \\
star &  stars, stellar, brightest, Stars, Galaxy, Stardust, eclipsing, stars., Star \\
cell & cells, DNA, cellular, cytoplasm, membrane, peptide, macrophages, suppressor, vesicles \\
left & leaving, turned, back, then, After, after, immediately, broke, end \\
\bottomrule
\end{tabular}
\end{small}
\end{center}
\caption{Nearest neighbors based on cosine similarity between the mean vectors of Gaussian components for Gaussian mixture embedding (top) (for $K=2$) and Gaussian embedding (bottom). The notation {\tt w:i} denotes  the $i^{th}$ mixture component of the word {\tt w}. 
}
\label{table:nn_2g}
\end{table*}

\section{Experiments} \label{section:experiments}

We have introduced a model for multi-prototype embeddings, which expressively captures word meanings with whole probability distributions.
We show that our combination of energy and objective functions, proposed in Section \ref{section:methodology}, enables one to learn interpretable multimodal distributions through unsupervised training, for describing words with multiple distinct meanings.  By representing multiple distinct meanings, our model also reduces the unnecessarily large variance of a Gaussian embedding model, and has improved results on word entailment tasks.

To learn the parameters of the proposed mixture model, we train on a
concatenation of two datasets: {\tt UKWAC} (2.5 billion tokens) and {\tt Wackypedia}  (1 billion tokens) \cite{wacky}. 
We discard words that occur fewer than $100$ times in the corpus, which results in a vocabulary size of $314,129$ words.  
Our word sampling scheme, described at the end of Section \ref{subsection:qualitative_eval}, is similar to that of {\tt word2vec} with one negative context word for each positive context word. 

After training, we obtain learned parameters $\{\vec{\mu}_{w,i}, \Sigma_{w,i}, p_i\}_{i=1}^K$ for each word $w$. 
We treat the mean vector $\vec{\mu}_{w,i}$ as the embedding of the $i^{\text{th}}$ mixture component  with the covariance matrix $\Sigma_{w,i}$ representing its subtlety and uncertainty. We perform qualitative evaluation to show that our embeddings learn meaningful multi-prototype representations and compare to existing models using a quantitative evaluation on word similarity datasets and word entailment.

We name our model as Word to Gaussian Mixture ({\tt w2gm}) in constrast to Word to Gaussian ({\tt w2g}) \cite{word2gauss}. Unless stated otherwise, {\tt w2g} refers to our implementation of {\tt w2gm} model with one mixture component. 

\subsection{Hyperparameters} \label{subsection:hyperparams}

Unless stated otherwise, we experiment with $K=2$ components for the {\tt w2gm} model, but we have results and discussion of $K=3$ at the end of section 4.3.  We primarily consider the spherical case for computational efficiency. We note that for diagonal or spherical covariances, the energy can be computed very efficiently since the matrix inversion would simply require $\mathcal{O}(d)$ computation instead of $\mathcal{O}(d^3)$ for a full matrix.  Empirically, we have found diagonal covariance matrices become roughly spherical after training.  Indeed, for these relatively high dimensional embeddings, there are sufficient degrees of freedom for the mean vectors to be learned such that the covariance matrices need not be asymmetric. Therefore, we perform all evaluations with spherical covariance models. 

Models used for evaluation have dimension $D=50$ and use context window $\ell = 10$ unless stated otherwise. We provide additional hyperparameters and training details in the supplementary material (\ref{sup:hyperparams}).

\subsection{Similarity Measures} \label{subsection:similarity_measures}

Since our word embeddings contain multiple vectors and uncertainty parameters per word, we use the following measures that generalizes  similarity scores. 
These measures  pick out the component pair with maximum similarity and therefore determine the meanings that are most relevant. 

\subsubsection{Expected Likelihood Kernel}
A natural choice for a similarity score is the expected likelihood kernel, an inner product between distributions, 
which we discussed in Section \ref{subsection:energy}.  This metric incorporates the uncertainty from the covariance 
matrices in addition to the similarity between the mean vectors.

\subsubsection{Maximum Cosine Similarity}
This metric measures the maximum similarity of mean vectors among all pairs of mixture components between distributions $f$ and $g$. That is, $\displaystyle d(f,g) = \max_{i,j= 1, \hdots, K} \frac{ \langle \bm{\mu}_{f,i}, \bm{\mu}_{g,j}  \rangle}{ ||\bm{\mu}_{f,i}|| \cdot || \bm{\mu}_{g,j} ||  }$, which corresponds to matching the meanings of $f$ and $g$ that are the most similar. For a Gaussian embedding, maximum similarity reduces to the usual cosine similarity.

\subsubsection{Minimum Euclidean Distance}
Cosine similarity is popular for evaluating embeddings. 
However, our training objective directly involves the Euclidean distance in Eq.~\eqref{eq:partial_energy}, as opposed to dot product of vectors such as in {\tt word2vec}. Therefore, we also consider the Euclidean metric: $\displaystyle d(f,g) = \min_{i,j= 1, \hdots, K} [  || \bm{\mu}_{f,i} -  \bm{\mu}_{g,j}  || ]  $.

\subsection{Qualitative Evaluation} \label{subsection:qualitative_eval}

In Table~\ref{table:nn_2g}, we show examples of polysemous words and their nearest neighbors in the embedding space to demonstrate that our trained  embeddings  capture multiple word senses. For instance, a word such as `rock' that could mean either `stone' or `rock music' should have each of its meanings represented by a distinct Gaussian component. Our results for a mixture of two Gaussians model confirm this hypothesis, where we observe that the $0^{th}$ component of `rock' being related to (`basalt', `boulders') and the $1^{st}$ component being related to (`indie', `funk', `hip-hop'). Similarly, the word {\tt bank} has its $0^{th}$ component representing the river bank and the $1^{st}$ component representing the financial bank.

By contrast, in Table~\ref{table:nn_2g} (bottom), see that for Gaussian embeddings with one mixture component,
nearest neighbors of polysemous words are predominantly related to a single meaning. For instance, `rock' mostly has  neighbors related to rock music and `bank' mostly related to the financial bank. The alternative meanings of these polysemous words are not well represented in the embeddings.  As a numerical example, the cosine similarity between `rock' and `stone' for the Gaussian representation of \citet{word2gauss} is only $0.029$, much lower than the cosine similarity $0.586$ between the $0^{th}$ component of `rock' and `stone' in our multimodal representation.

In cases where  a word only has a single popular meaning, the mixture components can be fairly close; for instance, one component of `stone' is close to (`stones', `stonework', `slab') and the other to (`carving, `relic', `excavated'), which reflects subtle variations in meanings. In general, the mixture can give properties such as heavy tails and more interesting unimodal characterizations of uncertainty than could be described by a single Gaussian.

\paragraph{Embedding Visualization}

We provide an interactive visualization as part of our code repository: \url{https://github.com/benathi/word2gm#visualization} that allows real-time queries of words' nearest neighbors (in the \texttt{embeddings} tab) for $K=1, 2, 3$ components. We use a notation similar to that of Table~\ref{table:nn_2g}, where a token {\tt w:i } represents the component {\tt i} of a word {\tt w}. For instance, if in the 
$K=2$ link we search for {\tt bank:0}, we obtain the nearest neighbors such as {\tt river:1, confluence:0, waterway:1}, which indicates that the $0^{\text{th}}$ component of `bank' has the meaning `river bank'. On the other hand, searching for {\tt bank:1} yields nearby words such as {\tt banking:1, banker:0, ATM:0}, indicating that this component is close to the `financial bank'.  We also have a visualization of a unimodal ({\tt w2g}) for comparison in the $K=1$ link.

In addition, the embedding link for our Gaussian mixture model with $K=3$ mixture components can learn 
three distinct meanings.  For instance, each of the three components of  `cell' is close to (`keypad', `digits'), (`incarcerated', `inmate') or (`tissue', `antibody'), indicating that the distribution captures the concept of `cellphone', `jail cell', or `biological cell', respectively.  Due to the limited number of words with more than $2$ meanings, our model with $K=3$ does not generally offer substantial performance differences to our model with $K=2$; hence, we do not further display $K=3$ results for compactness.


\begin{table*}[t]
\begin{center}
\begin{small}

\smallskip

\begin{tabular}{l|r|r | rrr | rrr}
\toprule
Dataset &    sg* & w2g* &  w2g/mc &  w2g/el &  w2g/me &  w2gm/mc &  w2gm/el  &  w2gm/me \\
\midrule
     SL & 29.39 &  \bf{32.23} &      \underline{29.35} &      25.44 &       25.43 &      \underline{29.31} &      26.02 &       27.59 \\
     {WS} & 59.89 &  65.49 &      \underline{71.53} &      61.51 &       64.04 &     \bf{\underline{ 73.47}} &      62.85 &       66.39 \\
   {WS-S} & 69.86 &  76.15 &       \underline{76.70} &      70.57 &        72.3 &      \bf{\underline{76.73}} &      70.08 &        73.3 \\
   {WS-R} & 53.03 &  58.96 &      \underline{68.34} &       54.4 &       55.43 &      \bf{\underline{71.75}} &      57.98 &       60.13 \\
    {MEN} & 70.27 &  71.31 &      \underline{72.58} &      67.81 &       65.53 &      \bf{\underline{73.55}} &       68.5 &        67.7 \\
     MC & 63.96 &  70.41 &      76.48 &       72.70 &       \bf{\underline{80.66}} &      79.08 &      76.75 &       \underline{80.33} \\
     {RG} & 70.01 &     71 &       \underline{73.30} &      72.29 &       72.12 &      \bf{\underline{74.51}} &      71.55 &       73.52 \\
     {YP} & 39.34 &   41.5 &      \underline{41.96} &      38.38 &       36.41 &      \bf{\underline{45.07}} &      39.18 &       38.58 \\
 {MT-287} &     - &      - &      \underline{64.79} &       57.5 &       58.31 &       \bf{\underline{66.60}} &      57.24 &       60.61 \\
 MT-771 &     - &      - &      \bf{\underline{60.86}} &      55.89 &       54.12 &      \underline{60.82} &      57.26 &       56.43 \\
     {RW} &     - &      - &      28.78 &      32.34 &       \underline{33.16} &      28.62 &      31.64 &       \bf{\underline{35.27}} \\
\bottomrule
\end{tabular}

\smallskip
\end{small}
\end{center}
\caption{Spearman correlation for word similarity datasets. 
The models {\tt sg, w2g, w2gm} denote {\tt word2vec} skip-gram, Gaussian embedding, and Gaussian mixture embedding (K=2). 
The measures {\tt mc, el, me} denote maximum cosine similarity, expected likelihood kernel, and minimum Euclidean distance. For each of {\tt w2g} and {\tt w2gm}, we underline the similarity metric with the best score.
For each dataset, we boldface the score with the best performance across all models.
%
The correlation scores for {\tt sg*, w2g*} are taken from \citet{word2gauss} and correspond to cosine distance.
}
\label{table:wordsimresults}
\end{table*}

\begin{table}[t]
\begin{center}
\begin{small}
\begin{sc}
\begin{tabular}{lrrrr}
\toprule
Model 			& 	$\rho \times 100$ \\
Huang		&	64.2 \\
Huang*		&	71.3 \\ 
MSSG 50d  & 63.2 \\ 
MSSG 300d  &  71.2 \\ 
w2g			& 70.9 \\ 
w2gm			&  \textbf{73.5} \\ 
\bottomrule
\end{tabular}
\smallskip
\end{sc}
\end{small}
\end{center}
\caption{Spearman's correlation ($\rho$) on WordSim-353 datasets for our Word to Gaussian Mixture embeddings, as well as the multi-prototype embedding by \citet{multipleprototypes} and the MSSG model by \citet{nonparam_multiprototype}. {\tt Huang*} is trained using data with all stop words removed. All models have dimension $D=50$ except for MSSG 300D with $D=300$ which is still outperformed by our {\tt w2gm} model.
}
\label{table:wordsim353_multi}
\end{table}

\subsection{Word Similarity} \label{subsection:wordsim}
We  evaluate our embeddings on several standard word similarity datasets, namely, SimLex \cite{evaldata_simlex}, WS or WordSim-353, WS-S (similarity), WS-R (relatedness) \cite{evaldata_wordsim}, MEN \cite{evaldata_men}, MC \cite{evaldata_mc}, RG \cite{evaldata_rg}, YP \cite{evaldata_yp}, MTurk(-287,-771) \cite{evaldata_mturk287, evaldata_mturk771}, and RW \cite{rarewords}. Each dataset contains a list of word pairs with a human score of how related or similar the two words are.

We calculate the Spearman correlation \cite{spearman04} between the labels and our scores generated by the embeddings. The Spearman correlation is a rank-based correlation measure that assesses how well the scores describe the true labels.

The correlation results are shown in Table~\ref{table:wordsimresults} using  the scores generated from the expected likelihood kernel, maximum cosine similarity, and maximum Euclidean distance.

We show the results of our Gaussian mixture model and compare the performance with that of {\tt word2vec} and the original Gaussian embedding by \citet{word2gauss}. We note that our model of a unimodal Gaussian embedding  {\tt w2g}  also outperforms the original model, which differs in model hyperparameters and initialization, for most datasets. 

Our multi-prototype model {\tt w2gm} also  performs  better than skip-gram or Gaussian embedding methods on many datasets, namely, {\tt WS, WS-R, MEN, MC, RG, YP, MT-287, RW}. The maximum cosine similarity yields the best performance on most datasets; however, the minimum Euclidean distance is a better metric for the datasets {\tt MC} and {\tt RW}. These results are consistent for both the single-prototype and the multi-prototype models.  

We also compare out results on WordSim-353 with the multi-prototype embedding method by \citet{multipleprototypes} and \citet{nonparam_multiprototype}, shown in Table~\ref{table:wordsim353_multi}. We observe that our single-prototype model {\tt w2g} is competitive compared to models by \citet{multipleprototypes}, even without using a corpus with stop words removed. This could be due to the auto-calibration of importance via the covariance learning which decrease the importance of very frequent words such as `the', `to', `a', etc. Moreover, our multi-prototype model substantially outperforms the model of \citet{multipleprototypes} and the MSSG model of \citet{nonparam_multiprototype} on the WordSim-353 dataset. 

\subsection{Word Similarity for Polysemous  Words} \label{subsection:wordsim2}

We use the dataset SCWS introduced by \citet{multipleprototypes}, where word pairs are chosen to have variations in meanings of polysemous and homonymous words. 

We compare our method with multiprototype models by {\tt Huang} \cite{multipleprototypes}, {\tt Tian} \cite{multi_word_embs},  {\tt Chen} \cite{unified_sense_chen14}, and {\tt MSSG} model by \cite{nonparam_multiprototype}. We note that {\tt Chen} model uses an external lexical source {\tt WordNet} that  gives it an extra advantage.

We use many metrics to calculate the scores for the Spearman correlation. {\tt MaxSim} refers to the maximum cosine similarity.  {\tt AveSim} is the average of cosine similarities with respect to the component probabilities.

In Table \ref{table:scws}, the model {\tt w2g} performs the best among all single-prototype models for either $50$ or $200$ vector dimensions.  Our model {\tt w2gm} performs competitively compared to other multi-prototype models.  In SCWS, the gain in flexibility in moving to a probability density approach appears to dominate over the effects of using a multi-prototype.  In most other examples, we see {\tt w2gm} surpass {\tt w2g}, where the multi-prototype structure is just as important for good performance as the probabilistic representation. 
Note that other models also use {\tt AvgSimC} metric which uses context information which can yield better correlation \cite{multipleprototypes, unified_sense_chen14}. We report the numbers using {\tt AvgSim} or {\tt MaxSim} from the existing models which are more comparable to our performance with {\tt MaxSim}.

\begin{table}[t]
\begin{center}
\begin{small}
\begin{sc}
\begin{tabular}{lrrrr}
\toprule
Model 			&  Dimension &	$\rho \times 100$ \\
word2vec skip-gram		&	50 & 61.7 \\
Huang-S		&	50 & 58.6 \\
w2g		&	50 & \bf{64.7} \\
Chen-S		&	200 & 64.2 \\
w2g		&	200 & \bf{66.2} \\
\midrule 
Huang-M AvgSim	&	50 & 62.8 \\
Tian-M MaxSim		&	50 & 63.6 \\
w2gm MaxSim	&  	50 & 62.7 \\ 
MSSG AvgSim & 50  & \bf{64.2} \\
Chen-M AvgSim	&	200 & \bf{66.2} \\ 
w2gm MaxSim	&  	200 & 65.5   \\ 
\bottomrule
\end{tabular}
\smallskip
\end{sc}
\end{small}
\end{center}
\caption{Spearman's correlation $\rho$ on dataset SCWS. We show the results for single prototype (top) and multi-prototype (bottom).
The suffix {\tt -(S,M)} refers to single and multiple prototype models, respectively.
 }
\label{table:scws}

\end{table}

\subsection{Reduction in Variance of Polysemous Words} \label{subsection:reduction_var}

One motivation for our Gaussian mixture embedding is to model word uncertainty more accurately than Gaussian embeddings, which can have overly large variances for polysemous words (in order to assign some mass to all of the distinct meanings).  We see that our Gaussian mixture model does indeed reduce the variances of each component for such words.  For instance, we observe that the word {\tt rock} in {\tt w2g} has much higher variance per dimension ($e^{-1.8} \approx 1.65 $) compared to that of  Gaussian components of {\tt rock} in {\tt w2gm} (which has variance of roughly $e^{-2.5} \approx 0.82$). We also see, in the next section, that {\tt w2gm} has desirable quantitative behavior for word entailment.

\subsection{Word Entailment} \label{subsection:entailment}
\begin{table}[t]
\begin{center}
\begin{small}
\begin{sc}
\begin{tabular}{c c c c c}
\toprule
Model & Score & Best AP	& Best F1 \\
w2g (5) 	& cos	& 	73.1		& 76.4  \\
w2g (5)	& kl		&	73.7		& 76.0 \\
w2gm (5)	& cos	& 	73.6		& 76.3 \\
w2gm (5)	& kl		&	75.7 		& 77.9 \\
\midrule
w2g (10)	& cos	& 73.0	&  76.1 \\
w2g (10)	& kl		& 74.2	& 76.1 \\
w2gm (10)	& cos	& 72.9	& 75.6 \\
w2gm (10)	& kl		&	74.7  & 76.3 \\
\bottomrule
\end{tabular}
\smallskip
\end{sc}
\end{small}
\end{center}
\caption{Entailment results for models {\tt w2g} and {\tt w2gm} with window size $5$ and $10$ for maximum cosine similarity and the maximum negative KL divergence. We calculate the best average precision and the best F1 score. In most cases, {\tt w2gm}  outperforms {\tt w2g} for describing entailment. 
}
\label{table:entailment}
\end{table}

We evaluate our embeddings on the word entailment dataset from \citet{baroni_entailment}. The lexical entailment between words is denoted by $w_1 \models w_2$ which means that all instances of  $w_1$ are $w_2$.   The entailment dataset contains positive  pairs such as \emph{aircraft} $\models $ \emph{vehicle} and negative pairs such as \emph{aircraft} $\not \models$ \emph{insect}.

We generate entailment scores of word pairs and find the best threshold, measured by Average Precision (AP) or F1 score, which identifies negative versus positive entailment. We use the maximum cosine similarity and the minimum KL divergence, $\displaystyle d(f,g) = \min_{i,j= 1, \hdots, K} KL(f || g)$, for entailment scores. The minimum KL divergence is similar to the maximum cosine similarity, but also incorporates the embedding uncertainty. In addition, KL divergence is an asymmetric measure, which is more suitable for certain tasks such as word entailment where a relationship is unidirectional. For instance, $w_1 \models w_2$ does not imply $w_2 \models w_1$.  Indeed, \emph{aircraft} $\models $ \emph{vehicle} does not imply \emph{vehicle} $\models $ \emph{aircraft}, since all aircraft are vehicles but not all vehicles are aircraft.  The difference between $KL(w_1 || w_2)$ versus $KL(w_2 || w_1)$ distinguishes which word distribution encompasses another distribution, as demonstrated in Figure~\ref{fig:multimodal}.

Table~\ref{table:entailment} shows the results of our {\tt w2gm} model versus the Gaussian embedding model {\tt w2g}. We observe a trend for both models with window size $5$ and $10$  that the  KL metric yields improvement (both AP and F1) over cosine similarity. 
In addition, {\tt w2gm} generally outperforms {\tt w2g}. 

The multi-prototype model estimates the meaning uncertainty better since it is no longer constrained to be unimodal, leading to better characterizations of entailment. On the other hand, the Gaussian embedding model suffers from overestimatating variances of polysemous words, which results in less informative word distributions and reduced entailment scores.


\section{Discussion} \label{section:discussion}

We introduced a model that represents words with expressive multimodal distributions formed from Gaussian mixtures.  To learn the properties of each mixture, we proposed an analytic energy function for combination with a maximum margin objective.  The resulting embeddings capture different semantics of polysemous words, uncertainty, and entailment, and also perform favorably on word similarity benchmarks.

Elsewhere, latent probabilistic representations are proving to be exceptionally valuable, able to capture nuances such as face angles with variational autoencoders \cite{autoencoding_vbayes} or subtleties in painting strokes with the InfoGAN \cite{infogan}.  Moreover, classically deterministic deep learning architectures are actively being generalized to probabilistic deep models, for full predictive distributions instead of point estimates, and significantly more expressive representations \citep{wilson2016deep, wilson2016stochastic, al2016learning, bayesrnn2016scalable, fortunato2017bayesian}.

Similarly, probabilistic word embeddings can capture a range of subtle meanings, and advance the state of the art.  Multimodal word distributions naturally represent our belief that words do not have single precise meanings: indeed, the shape of a word distribution can express much more semantic information than any point representation.  

In the future, multimodal word distributions could open the doors to a new suite of applications in language modelling, where whole word distributions are used as inputs to new probabilistic LSTMs, or in decision functions where uncertainty matters.  As part of this effort, we can explore different metrics between distributions, such as KL divergences, which would be a natural choice for order embeddings that model entailment properties.  It would also be informative to explore inference over the number of components in mixture models for word distributions.  Such an approach could potentially discover an unbounded number of distinct meanings for words, but also distribute the support of each word distribution to express highly nuanced meanings.  Alternatively, we could imagine a dependent mixture model where the distributions over words are evolving with time and other covariates.  One could also build new types of supervised language models, constructed to more fully leverage the rich information provided by word distributions.
 
\subsection*{Acknowledgements}
We thank NSF IIS-1563887 for support.

\footnotesize{
\bibliography{multimodal_word_embeddings}
\bibliographystyle{acl17-latex/acl_natbib}
}

\clearpage
\appendix
\label{sec:supplementary}

\section{Supplementary Material} \label{section:supplementary}

\subsection{Derivation of Expected Likelihood Kernel}  \label{subsection:derivation}
We derive the form of expected likelihood kernel for Gaussian mixtures. Let $f,g$ be Gaussian mixture distributions representing the words $w_f, w_g$. That is,  $f(x) = \sum_{i=1}^K p_i \Nor(x; \mu_{f,i} , \Sigma_{f,i} ) $ and $g(x)  =  \sum_{i=1}^K q_i \Nor(x; \mu_{g,i} , \Sigma_{g,i} )$, $\sum_{i =1}^K p_i = 1 $, and $\sum_{i =1}^K q_i = 1$. The expected likelihood kernel is given by 
\begin{align*}
E_\theta(f,g) 	&=	\int  \left( \sum_{i=1}^K p_i \Nor(x; \mu_{f,i} , \Sigma_{f,i} ) \right) \cdot \\  & \left( \sum_{j=1}^K q_j \Nor(x; \mu_{g,j} , \Sigma_{g,j} ) \right) \ d x \\ 
&= \sum_{i=1}^K \sum_{j=1}^K  p_i q_j \int  \Nor(x; \mu_{f,i} , \Sigma_{f,i} ) \cdot \Nor(x; \mu_{g,j} , \Sigma_{g,j} ) \ d x \\ 
&= \sum_{i=1}^K \sum_{j=1}^K  p_i q_j \Nor(0; \mu_{f,i} - \mu_{g,j} , \Sigma_{f,i} + \Sigma_{g,j} ) \\
&= \sum_{i=1}^K \sum_{j=1}^K p_i q_j  e^{\xi_{i,j}} 
\end{align*}
where we note that $\int \Nor(x; \mu_i, \Sigma_i) \Nor(x; \mu_j, \Sigma_j) \ dx = \Nor(0, \mu_i - \mu_j , \Sigma_i + \Sigma_j)$ \cite{word2gauss} and $\xi_{i,j}$ is the log partial energy, given by equation ~\ref{eq:partial_energy}.

\subsection{Implementation}
\label{sec:implementation}
In this section we discuss practical details for training the proposed model.

\subsubsection*{Reduction to Diagonal Covariance}

We use a diagonal $\Sigma$, in which case inverting the covariance matrix is trivial and computations are particularly efficient.

 Let $\bm{d}^f, \bm{d}^g$ denote the diagonal vectors of $\Sigma_f, \Sigma_g$ The expression for $\xi_{i,j}$ reduces to \begin{align*}
\xi_{i,j}
= 	- \frac{1}{2} \sum_{r=1}^D \log (  d^p_r + d^q_r) 
\\       - \frac{1}{2} \sum \left[ (\bm{\mu}_{p,i} - \bm{\mu}_{q,j})  \circ \frac{1}{ \bm{d^p + d^q}   } \circ (\bm{\mu}_{p, i} - \bm{\mu}_{q,j}) \right]
\end{align*}
where $\circ$ denotes element-wise multiplication. The spherical case which we use in all our experiments is similar since we simply replace a vector $\bm{d}$ with a single value.

\subsubsection*{Optimization Constraint and Stability}

We optimize  $\log \bm{d}$ since each component of diagonal vector $\bm{d}$ is constrained to be positive. Similarly, we constrain the probability $p_i$ to be in $[0,1]$ and sum to $1$ by optimizing over unconstrained scores $s_i \in (-\infty, \infty)$ and using a softmax function to convert the scores to probability $p_i = \frac{e^{s_i}}{\sum_{j=1}^K e^{s_j} }$. 

The loss computation can be numerically unstable if elements of the diagonal covariances are very small, due to the term $ \log (  d^f_r + d^g_r) $ and $ \frac{1}{ \bm{d}^q + \bm{d}^p} $. Therefore, we add a small constant $\epsilon = 10^{-4}$ so that $d^f_r + d^g_r$ and $ \bm{d}^q + \bm{d}^p $ becomes $d^f_r + d^g_r + \epsilon$ and $ \bm{d^q + d^p} + \epsilon$.

In addition, we observe that $\xi_{i,j}$ can be very small which would result in $e^{\xi_{i,j}} \approx 0$ up to machine precision. In order to stabilize the computation in eq.~\ref{eq:loge}, we compute its equivalent form
\[
\log E(f,g) = \xi_{i',j'} +  \log \sum_{j=1}^K \sum_{i=1}^K p_i q_j e^{\xi_{i,j} - \xi_{i',j'}}
\]
where $ \xi_{i',j'} = \max_{i,j} \xi_{i,j}$.

\subsubsection*{Model Hyperparameters and Training Details} \label{sup:hyperparams}

 In the loss function $L_\theta$, we use a margin $m= 1$ and a batch size of $128$.  We initialize the word embeddings with a uniform distribution over $[ -\sqrt{\frac{3}{D}}, \sqrt{\frac{3}{D}} ]$ so that the expectation of variance is $1$ and the mean is zero \cite{lecun-efficient-backprop}. We initialize each dimension of the diagonal matrix (or a single value for spherical case) with a constant value $v = 0.05$. We also initialize the mixture scores $s_i$ to be $0$ so that the initial probabilities are equal among all $K$ components. 
 We use the threshold $t = 10^{-5}$ for negative sampling, which is the recommended value for {\tt word2vec} skip-gram on large datasets.

We also use a separate output embeddings in addition to input embeddings, similar to {\tt word2vec} implementation \cite{word2vec2, word2vec1}. That is, each word has two sets of distributions $q_{I}$ and $q_{O}$, each of which is a Gaussian mixture. For a given pair of word and context $(w,c)$, we use the input distribution $q_{I}$ for $w$ (input word) and the output distribution $q_{O}$ for context $c$ (output word). We optimize the parameters of both $q_{I}$ and $q_{O}$ and use the trained input distributions $q_{I}$ as our final word representations.

We use mini-batch asynchronous gradient descent with Adagrad \cite{adagrad} which performs adaptive learning rate for each parameter. We also experiment with Adam \cite{adam} which corrects the bias in adaptive gradient update  of Adagrad and is proven very popular for most recent neural network models. However, we found that it is much slower  than Adagrad ($\approx 10$ times). This is because the gradient computation of the model is relatively fast, so a complex gradient update algorithm such as Adam becomes the bottleneck in the optimization. Therefore, we choose to use Adagrad which allows us to better scale to large datasets. We use a linearly decreasing learning rate from $0.05$ to $0.00001$. 

\end{document}